\documentclass[conference]{IEEEtran}
\IEEEoverridecommandlockouts
\usepackage{cite}
\usepackage{amsmath,amssymb,amsfonts}
\usepackage{algorithm}
\usepackage{algorithmic}
\usepackage{graphicx}
\usepackage{multirow}
\usepackage{diagbox}
\usepackage{textcomp}
\usepackage{xcolor}
\def\BibTeX{{\rm B\kern-.05em{\sc i\kern-.025em b}\kern-.08em
    T\kern-.1667em\lower.7ex\hbox{E}\kern-.125emX}}
\begin{document}

\title{Event Temporal Relation Extraction based on Retrieval-Augmented on LLMs  \\
%{\footnotesize \textsuperscript{*}Note: Sub-titles are not captured in Xplore and should not be used}
%\thanks{Identify applicable funding agency here. If none, delete this.}
}

\author{\IEEEauthorblockN{Anonymous Authors}}
\iffalse
\author{\IEEEauthorblockN{1\textsuperscript{st} Given Name Surname}
\IEEEauthorblockA{\textit{dept. name of organization (of Aff.)} \\
\textit{name of organization (of Aff.)}\\
City, Country \\
email address or ORCID}
\and
\IEEEauthorblockN{2\textsuperscript{nd} Given Name Surname}
\IEEEauthorblockA{\textit{dept. name of organization (of Aff.)} \\
\textit{name of organization (of Aff.)}\\
City, Country \\
email address or ORCID}
\and
\IEEEauthorblockN{3\textsuperscript{rd} Given Name Surname}
\IEEEauthorblockA{\textit{dept. name of organization (of Aff.)} \\
\textit{name of organization (of Aff.)}\\
City, Country \\
email address or ORCID}
\and
\IEEEauthorblockN{4\textsuperscript{th} Given Name Surname}
\IEEEauthorblockA{\textit{dept. name of organization (of Aff.)} \\
\textit{name of organization (of Aff.)}\\
City, Country \\
email address or ORCID}
\and
\IEEEauthorblockN{5\textsuperscript{th} Given Name Surname}
\IEEEauthorblockA{\textit{dept. name of organization (of Aff.)} \\
\textit{name of organization (of Aff.)}\\
City, Country \\
email address or ORCID}
\and
\IEEEauthorblockN{6\textsuperscript{th} Given Name Surname}
\IEEEauthorblockA{\textit{dept. name of organization (of Aff.)} \\
\textit{name of organization (of Aff.)}\\
City, Country \\
email address or ORCID}
}
\fi
\author{\IEEEauthorblockN{Xiaobin Zhang, Liangjun Zang$^{\ast}$\thanks{*Corresponding author}, Qianwen Liu, Shuchong Wei, Songlin Hu}
\IEEEauthorblockA{
\textit{Institute of Information Engineering, Chinese Academy of Sciences. Beijing, China}\\
\textit{School of Cyber Security, University of Chinese Academy of Sciences. Beijing, China} \\
}}

\maketitle

\begin{abstract}
Event temporal relation (TempRel) is a primary subject of the event relation extraction task. However, the inherent ambiguity of TempRel increases the difficulty of the task. With the rise of prompt engineering, it is important to design effective prompt templates and verbalizers to extract relevant knowledge. The traditional manually designed templates struggle to extract precise temporal knowledge. This paper introduces a novel retrieval-augmented TempRel extraction approach, leveraging knowledge retrieved from large language models (LLMs) to enhance prompt templates and verbalizers. Our method capitalizes on the diverse capabilities of various LLMs to generate a wide array of ideas for template and verbalizer design. Our proposed method fully exploits the potential of LLMs for generation tasks and contributes more knowledge to our design. Empirical evaluations across three widely recognized datasets demonstrate the efficacy of our method in improving the performance of event temporal relation extraction tasks.

\end{abstract}


\section{Introduction}
%%%%%%%%%%%%%%%%%%%%%%%%%%%%%%%%%%%%%%%%%%%%%%%%%%%%%%%%%%%%%%%%%%%
%背景与问题
Event temporal relation (TempRel) is a crucial task for Event-Related tasks. It can facilitate various downstream tasks (e.g., information retrieval, event timeline generation, summarization, etc.). However, the TempRel tasks also face many challenges, one of the most difficult problems is the ambiguity of temporal relation, that is, it is difficult to accurately capture the temporal relationship between events from the context information, which leads to difficulty in extracting temporal relationship. This makes TempRel tasks very difficult and hard to improve its performance,
%引出研究点
to solve this problem, in addition to the widely used data augmentation methods, prompt-based methods have also become a hot research topic in recent years, the key to designing a good prompt is to design a good pair of PVP\footnote{In this paper, we adopt the term PVP (pattern-verbalizer pair) where ‘pattern’ is an alias for ‘template’.}, especially the template design, it is becoming very important to extract the desired knowledge from the language model (LM). Therefore, it is vital to design a good prompt template for the TempRel task.

%我们的研究基础
Taking template design as an example, we tried to let the large model generate the TempRel template, but the performance was not as expected, so we narrowed down the search scope and started with the partial template design. Following the PTR\cite{han2021ptr} model, we expanded on it by manually designing three types of prompt templates(Post-style, Pre-style, and QA Style), and testing them separately. 
%问题
As shown in Table \ref{preliminaryExp}, we find two conclusions: First, the Post-decorated(\#1) templates outperformed the pre-decorated(\#4) and QA style(\#5) templates at the same pair of modifiers(e.g., $\langle$Head, Tail$\rangle$). Second, in the same style, the results of different modifiers are significantly different(e.g., \#1, \#2 and \#3). Therefore, it is very important to find an appropriate trigger word modifier for the event temporal relation template. The widely used manual design method based on empirical experience cannot meet the needs. The same problem exists in the design of the prompt verbalizer.

With the rapid development of Large Language Models(LLMs), the Retrieval-Augmented Generation(RAG) techniques have been widely used in various downstream tasks. In QA and dialogue situations, we can get answers through retrieval and generation. The retrieval-augmented generation core idea is a mixture framework that integrates the retrieval model and generation model, which generates text that is not only accurate but also rich in information. The RAG core idea is to first collect sufficient evidence related to the question, and then find the answer by the Large Language Model. Therefore, for template design, our research focuses on how to use LLMs to select appropriate modifiers for the trigger words. In the design of verbalizers, our main focus is on how to utilize large language models (LLMs) to establish an appropriate mapping from the vocabulary space to the label space.

\setlength{\tabcolsep}{1mm}{
% Please add the following required packages to your document preamble:
% \usepackage{multirow}

\begin{table}[]
\caption{The performances of different style templates on TB-Dense We use the same soft verbalizer WARP \cite{warp} for all templates.}\label{preliminaryExp}
\centering
\begin{tabular}{|c|l|c|}
\hline
Style                & \multicolumn{1}{c|}{Template}                            & F1    \\ \hline
\multirow{3}{*}{Post} & \#1:\textbf{e1} (Head) {[}MASK{]} \textbf{e2} (Tail)                       & 66.96 \\ \cline{2-3} 
                      & \#2:\textbf{e1} (Event) {[}MASK{]} \textbf{e2} (Event)                     & 67.83 \\ \cline{2-3} 
                      & \#3:\textbf{e1} (Trigger) {[}MASK{]} \textbf{e2} (Trigger)                 & 68.33 \\ \hline
Pre                   & \#4: It is Head \textbf{e1} {[}MASK{]} It is Tail \textbf{e2}              & 64.82 \\ \hline
QA                    & \#5:\textbf{e1} is Head and \textbf{e2} is Tail,the relation is {[}MASK{]} & 62.86 \\ \hline
\end{tabular}
\end{table}

}

%我们的方法
In this paper, we propose a novel RAG-based model for Event Temporal Relation(RETR), which is a prompt-based neural network method that focuses on addressing the task of the TempRel framework. The RETR consists of two parts, the rough selection stage and the fine-tuning selection stage. In the first stage, we list various strategies with prompts in terms of PLMs selection, template style design, and tuning modes. In the second stage, our goal is to search for suitable PVP pairings, which utilize the Retrieval-Augmented Generation techniques for the prompt template to optimize the trigger word modifier word and find the suitable verbalizer for each TempRel dataset using the LLMs. In the phase of searching for suitable PVP pairings, we designed an algorithm to find the best PVP pairings based on the optimal strategy of RETR. After the above two stages, we can obtain the best extraction performance and get the best F1 value. Experimental results show that our method consistently achieves good performances on three widely used TempRel datasets. Besides its effectiveness, the model is universal, which is valuable for future related applications.%贡献
 Our contributions can be summarized as follows:
\begin{itemize}

    \item We are the first to integrate Retrieval-Augmented Generation (RAG) with the prompt-based learning paradigm, mitigating the problem of relation ambiguity in the task of event TempRel extraction.

    \item We propose a novel method that utilizes Retrieval-Augmentation techniques to fully mine inherent TempRel knowledge from various LLMs. This method leverages the distinct characteristics of different LLMs to help us design and find the most appropriate PVPs.

    \item Experimental results have shown that our method achieved outstanding results on three widely-used TempRel datasets \emph{TB-Dense},\emph{TDD-Man}, and \emph{TDD-Auto}.
\end{itemize}

\section{Related work}

\subsection{TempRel Extraction}

Early temporal relation extraction is based on rule-based methods, which utilize classic machine learning methods such as Perceptron and SVM. A notable work is CAEVO\cite{Chambers14denseevent}, which employs refined linguistic and syntactic rules, resulting in superior performance compared to previous work. Various deep-learning approaches have been proposed to address TempRel tasks. 
Ning et al.\cite{ning-etal-2019-improved} uses an LSTM encoder combined with the temporal prior knowledge\cite{ning-etal-2018-improving} to help extract temporal relation. Cheng et al.\cite{cheng-miyao-2017-classifying} implemented a bidirectional LSTM model, which is based on dependency paths for event temporal relation extraction without additional resources. EventPlus\cite{ma-etal-2021-eventplus} is a temporal event pipeline model that seamlessly integrates several state-of-the-art event understanding components. which combines event duration detection and temporal relation extraction, making it a highly effective tool for users seeking to quickly acquire event and temporal information. Mathur et al.\cite{mathur-etal-2021-timers} devised a gated GCN that incorporates syntactic elements, discourse markers, and time-based features into semantic roles for document-level TempRel extraction. Leveraging a gated Graph Convolutional Network (GCN), they were able to effectively capture the intricate relationships between entities and their temporal features. Zhou et al.\cite{Zhou2021ClinicalTR} has proposed a probabilistic soft logic regularization(PSL) approach, based on global inference for document-level TempRel extraction. Han et al.\cite{han-etal-2021-econet} developed a continuous training framework that integrates target mask prediction and contrastive loss, enabling the PLM to effectively mine the knowledge of temporal relationships. Man et al.\cite{TrongSOCS} utilized a reinforcement learning approach to identify pivotal sentences within documents for TempRel extraction, which enhanced the accuracy and efficiency of temporal relation extraction. Zhang et al.\cite{KJETE} employed temporal commonsense knowledge to augmente data, while also balancing the labels for more accurate joint event-TempRel extraction. This innovative strategy not only augments the data quality but also addresses the class imbalance problem, enabling the robust and reliable extraction of temporal relations. Wang et al.\cite{DBLP:conf/iconip/WangYM23} proposed a two-stage graph convolutional network that utilizes node representation, dependency type representation, and dependency type weight to optimize the node representation in the first stage. Subsequently, it generates the adjacency matrix based on the dependency tree for relation extraction. Huang et al.\cite{huang-etal-2023-classification} proposed a unified framework that offers a comprehensive perspective, considering temporal relations as the start and end time points of two events. This unified approach provides a structured framework for analyzing and representing temporal relationships.

\subsection{Prompt-based Learning}
In recent years, prompt-based methods approaches have been proven to be effective in improving the performance of TempRel extraction by leveraging the knowledge inherited in PLM. AdaPrompt\cite{Chen2021KnowPromptKP}  was a pioneer in the use of fixed-prompt language model tuning for relation extraction. It introduced an adaptive mechanism that distributed label words among different character numbers, enabling it to accommodate multiple label spaces. This approach allowed for more flexibility and adaptability in handling different types of label spaces, thus improving the overall performance of relation extraction. Subsequently, the PTR\cite{han2021ptr} model developed a set of sub-prompts that were later integrated into specific task prompts using logical rules. This integration ensured that prior knowledge was seamlessly incorporated into the prompt template, enhancing their overall performance. The KPT\cite{hu-etal-2022-knowledgeable} model expands the verbalizer in prompt-tuning by leveraging external knowledge base. Besides, the LM-BFF\cite{gao-etalLMBFF} model employed a prompt-based fine-tuning strategy, which automatically generates templates using the T5 model. This innovative approach not only improves the diversity of templates created but also enables the model to quickly and efficiently search for the best template. With the emergence of ChatGPT, Chan et al.\cite{Chan2023ChatGPTEO} proposed two prompt-based strategies to trigger the TempRel extraction ability of the large language model: one utilized an existing prompt template from Robinson et al.,\cite{Robinson2022LeveragingLL} and the other method manually designed a more sophisticated prompt template.

\subsection{Retrieval-Augmented Generation}
%基础RAG流程简单来说如下：将文本分割成块，然后使用编码模型将这些块嵌入到向量中，将所有这些向量放入索引中，最后为LLM创建一个提示，告诉模型根据我们在搜索步骤中找到的上下文来回答用户的查询。
%The basic idea of the Retrieval-Augmented Generation(RAG) process can be summarized as follows: the text is first segmented into blocks, which are then embedded into vectors using an encoding model. Put all those vectors into an index, and finally create a prompt for the LLM that tells the model to answer the user's query based on the context we found during the search step.
RAG is a widely used technique that relies on Large Language Models (LLMs) for its implementation. This method effectively incorporates search algorithm-derived information into the context and integrates both the query and retrieved context into the prompt sent to the LLM. This innovative approach enables the model to provide more relevant and high-quality responses to queries.

Lewis et al.\cite{Lewis2020RetrievalAugmentedGF} explore a RAG method that combines pre-trained parametric and non-parametric memories for language generation. Jiang et al.\cite{jiang-etal-2023-structgpt} proposed an Iterative Reading-then-Reasoning (IRR) framework to solve question-answering tasks based on the structured data. ReAugKD\cite{zhang-etal-2023-reaugkd} introduces a novel framework and loss function that preserves the semantic similarities between teacher and student training examples. It ensures that the student effectively leverages the knowledge base to improve its performance and generalization capabilities. Xu et al.\cite{xu-etal-2023-retrieval} propose a two-level selection scheme to filter out irrelevant noise and extract the most relevant information. This involves a differentiable sampling module and an attention mechanism for passage-level and word-level selection. Yu et al.\cite{yu-etal-2023-retrieval} introduce a non-trivial method, which has two innovative training objectives: EM-L and R-L. These objectives provide more task-specific guidance to the retrieval metric, utilizing the EM algorithm and a ranking-based loss, respectively. This approach is suitable for few-shot scenarios.

\section{TempRel Task}
\textbf{Task Description} Given a document $D$, it consists of a set of sentences $D=\{s_{1},s_{2},...,s_{n}\}$ and a set of events $\mathcal{E}=\{e_{1},e_{2},...,e_{m}\}$. Our temporal relation extraction aims at extracting the event temporal relations $r(e_{i},e_{j})$, where $e_{i},e_{j} \in \mathcal{E}$, $r(e_{i},e_{j}) \in {R}_{\tau}$ and ${R}_{\tau} $ is a pre-defined set of all possible TempRel labels.

\begin{figure}[!htbp]
  \centering
  \includegraphics[width=.47\textwidth]{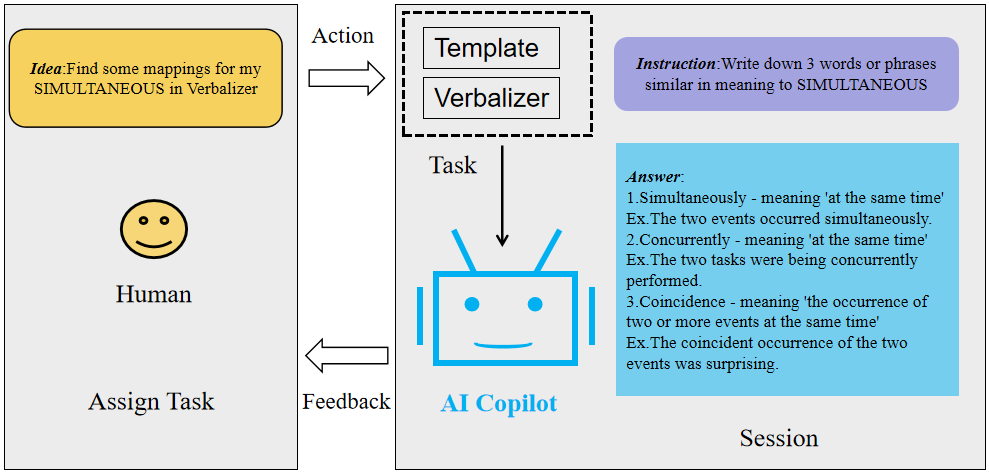}
  \caption{The overall architecture of our model.}
  \label{tablerag}
\end{figure}

\subsection{Overrall Architecture}
%We use the following Large Language Model as our auxiliary Retrieval-Augmented tool: Ernie-Bot3.5, Baichuan, ChatGLM, and LLama-2\cite{Touvron2023Llama2O}. We obtain their answers with QA prompt as the multi-turn dialogue history. 
Our method is depicted in Figure \ref{tablerag}. To fully unleash the potential capabilities of prompt templates, we aim to improve the reasoning ability of large language models (LLMs) over structured data in a unified way. Inspired by the studies on tool augmentation for LLMs, we developed an Iterative Question-then-Answering (IQA) framework to solve tasks based on the TempRel dataset, called \textbf{PETR}. In this framework, we manually designed three types of template styles and did the first round of roughing, then, in the second round, we used several LLMs to select appropriate modifiers for the templates. By iterating this procedure with the provided question, our approach can gradually approach the target answers to a given query. 

\subsection{Fine-tuning PLM for RAG}
% Specifically, the process of finding the best PVP, the model network can be expressed as $f$ formulation can be written as:
The optimal PVP pairing is obtained by selecting the appropriate PVP pairing on the dataset. The process of finding the best PVP can be formulated as:

\begin{equation}
    \begin{aligned}\label{zlj}
      & T^*(x),v^*(y) = \\
      & \mathop{\arg\max}_{T(x),v(y)\in \psi_e} { \mathop{\boldsymbol{\Sigma}}_{i=1}^{n}{\log p([\text{MASK}]=v(y_i)|[\boldsymbol{x}_i;T(\boldsymbol{x}_i)]);\boldsymbol{\theta}))} } \\
    \end{aligned}
\end{equation}
Where the PVP consists of $T(x)$ and $v(y)$, $\mathcal{\psi}_{e}$ is the set of candidate PVPs.  $\boldsymbol{x}$ represents the input sentence, and $[$MASK$]$ represents the concrete temporal words to be predicted by the PLM. $\theta$ denotes the model's parameters.  The crucial part of the template design is to fill the modifier words in the brackets. $(T^{*}(x), v^{*}(y))$ denotes the PVP with the best score. This is shown in Table \ref{tablesearchscheme}.

\setlength{\tabcolsep}{2.5mm}{
    % Please add the following required packages to your document preamble:
% \usepackage{multirow}
\begin{table}[!htbp]
\centering
\caption{Example of PVPs $\mathcal{\psi}_{e}$ composition.}\label{tablesearchscheme}
\begin{tabular}{c|c|c}
\hline
\multirow{2}{*}{No} & P(x)                                     & v(l)      \\ \cline{2-3} 
                    & Modifier                                 & Mode      \\ \hline
1                   & \textless{}Head,Tail\textgreater{}       & hard/soft \\ \hline
2                   & \textless{}Event,Event\textgreater{}     & hard/soft \\ \hline
3                   & \textless{}Pivotal,Pivotal\textgreater{} & hard/soft \\ \hline
4                   & ......                                   & ......    \\ \hline
\end{tabular}
\end{table}
}

Accordingly, we proposed an algorithm to find the most suitable PVP pairing. The process above can be illustrated by Algorithm \ref{alg:algorithm}.

\begin{algorithm}[]
\caption{PVP Instruction Finetuning}
\label{alg:algorithm}
\textbf{Input}: The finetuning model f($\cdot$),
event temporal relation dataset  $\mathcal{D}_{e}$ \\
\textbf{Output}:The best PVP pairing $\hat{\eta}$ \\
\begin{algorithmic}[1] %[1] enables line numbers

%\emph{(1).Select the best PVP pair on the test dataset}
\STATE Create an initial pairs $\mathcal{\psi}_{e}$. \\
\STATE Design $\{(p(x),v(l))_{i}\}_{i=1,...,N}$ pvp pairs manually based on Wikipedia, add them to $\mathcal{\psi}_{e}$. \\
\emph{(1). Expand candidate pairs with RAG}
\STATE Design prompts to query the candidate LLMs and get the $\{(p(x),v(l))_{i}\}_{i=1,..., M}$ pvp pair returned by the corresponding LLMs, add them to $\mathcal{\psi}_{e}$.

\emph{(2).Select the best PVP pair on dataset}

\STATE Create an initial best pvp pair $\hat{\eta}$
\STATE Get the N+M PVP pairs from the first phase $\mathcal{\psi}_{e}$
\STATE Set $best\_value$ $\xleftarrow{}$ 0.

\FORALL{$(p(x)^{+}_e, v(l)^{+}_{e}) \in \mathcal{\psi}_{e}$}
\STATE Verify the F1 value on the validation dataset $\mathcal{D}_{e}$, $\gamma^{+} \xleftarrow{}$  $f(p(x)^{+}_e, v(l)^{+}_{e}) $

\IF {$ \gamma^{+} \ge best\_value $}
    
\STATE $best\_value$ $\xleftarrow{}$  $\gamma^{+}$.
\STATE  $\hat{\eta}$ $\xleftarrow{}$ $\langle p(x)^{+}_e, v(l)^{+}_{e})\rangle $
\ELSE
\STATE abandoned the $(p(x)^{+}_e, v(l)^{+}_{e})$.
\ENDIF
\ENDFOR

\STATE \textbf{return} the best pvp pairing $\hat{\eta}$.
\end{algorithmic}
\end{algorithm}

Accordingly, we proposed an algorithm to find the most suitable PVP($T^{*}(x), v^{*}(y)$). First, we designed several trigger modifiers based on Wikipedia. Secondly, we wrote prompts to query several widely used LLMs to obtain the template modifier words and fill these modifier words into the Modifier bracket in Table \ref{tablesearchscheme}. Third, we query the LLMs to generate the mapping words to the individual labels in the verbalizer and treat them as the hard type in Table \ref{tablesearchscheme}. Fourth, we combined the modifier words with hard or soft verbalizers to form a PVP. Then, we applied these PVPs on the validation dataset to select the one that works the best. 

\subsection{PVP training and Reasoning}
Given a sequence of input events message $x$, we define the model label $y \epsilon \zeta $, the objective of the fine-tuning and optimization process is as follows:

\begin{equation}
    \begin{aligned}\label{pred1}
          s(y|x) &= M(y|(x,<P(x),v(l)>);\theta) \\ 
    \end{aligned}    
\end{equation}

Where $P(x)$ is the model's template, $v(l)$ is the model's verbalizer. $M$ represents the model. $s$ is the logits prediction by the model. $\theta$ is the parameter of the model. Then, we can get the predicted token $p(y|x)$ with a maximum probability calculated by the \textrm{softmax}. 

%Furthermore, we address the problem of fine-tuning PVP pairing with cross-entropy loss. 
\begin{equation}
    \begin{aligned}\label{pred2}
        &  p(y|x) = \frac{e^{s(y|x)}}{\sum_{s^{'}\epsilon \zeta}e^{s^{'}(y|x)}}
    \end{aligned}    
\end{equation}
After this step, we will obtain the final event temporal relation ${R}_{\tau} $, where the ${R}_{\tau} $ have been defined in the preceding paragraph. 

Specifically, the way to find the best PVP pairing $\hat{\eta}$ formulation can be written as:
\begin{equation}
    \begin{aligned}\label{predsearch}
      & \hat{\eta} =  \underset{(P(x),v(l)) \epsilon \mathcal{\psi}_{e}}{\arg\max} f_{prompt}(x,(P(x),v(l));\theta) \\
    \end{aligned}
\end{equation}

Where $\theta$ denotes the model's parameters, $\mathcal{\psi}_{e}$ is the candidate set of PVP pairs. $\hat{\eta}$ denotes the best score of the PVP scheme.
%First, we designed some combination of methods described as Eq.(\ref{tablesearchscheme}), and then combined hard and soft verbalizers to form a PVP pairing. 

%%%%%%%%%%%%%%%%%%%%%%%%%%%%%%%%%%%%%%%%%%%%%%%%%%%%%%%%%%%%%%%%%%%
\section{Experiments}
In this section, we first present the datasets we used, the details of our implementation, the model tuning approach, and the benchmark models against which we compared our performance. To verify the effectiveness of our model, We delve into the overall performance metrics as well as the individual label performance breakdowns. In addition, we also demonstrate multi-group comparison experiments to verify the validity of our model in detail.

\subsection{Datasets}
We conducted experiments on the widely utilized event temporal relation datasets:
TB-Dense\cite{cassidy-EtAl:2014:P14-2} and TDDiscourse\cite{naik-etal-2019-tddiscourse}. Where TB-Dense is derived from the TimeBank News Corpus \cite{2003TB}. It comprises local event pairs, indicating that events with temporal relation are found within the same sentence or adjacent sentences. The TDDiscourse is the event temporal relation dataset that first focuses on discourse-level TempRel, annotating temporal relationships between events that are separated by multiple sentences. It comprises two subsets: TDD-Man and TDD-Auto. All of these datasets above adopt the TimeML-based annotations that utilize TLINKs (temporal links) to express temporal relationships between two events. A statistical analysis of the three datasets is presented in Table \ref{tableds}.

\setlength{\tabcolsep}{2.5mm}{
%[!htbp]
\begin{table}[!htbp]
\centering
\caption{A statistical analysis of the TB-Dense, and the \\ TDDiscourse datasets (only including event-event TLINKs).}\label{tableds}
\begin{tabular}{|c|c|c|c|}
\hline
\diagbox{Dataset}{Split} & Train & Valid & Test \\ \hline\
%Documents      & 22    & 5   & 9    \\ \hline
TB-Dense & 4032  & 629 & 1427 \\ \hline
TDD-Man & 4000  & 650 & 1500 \\ \hline
TDD-Auto & 32609  & 1435 & 4258 \\ \hline
\end{tabular}
\end{table}
}

%%%%%%%%%%%%%%%%%%%%%%%%%%%%%%%%%%%%%%%%%%%%%%%%%%%%%%%%%%%%%%%%%%%
\subsection{Implementation Details}
In the experiment, we implemented the model based on the OpenPrompt\footnote{The open-source project address is https://github.com/thunlp/OpenPrompt and the official documentation address is https://thunlp.github.io/OpenPrompt/} framework, and designed templates and verbalizers style according to the characteristics of the TempRel task. We use the eight auxiliary LLMs(i.e., LLaMa-2\cite{Touvron2023Llama2O}, GPT 4.0, Ernie-Bot3.5, Baichuan, Taichu\footnote{https://taichu-web.ia.ac.cn/}, ChatGLM2-6B, Qwen-7B, and SparkDesk) to implement RAG.
We carried out experiments on four NVIDIA Tesla V100 GPUs and evaluated the performance of our model using the micro-F1 score. We choose RoBERTa-Large as the Pre-trained Language Model to encode the input of the event corpus. Through the hyper-parameter search in the experiment, we found the best performance can be achieved with the following settings: a maximum length of 112, a batch size of 16, total training epochs is 200, a learning rate of 5e-5, and a hidden size of 100. 

\subsection{Overall Performance}
%We carried out the experiments summarized as follows:
The overall results are summarized in Table \ref{mainresult}.

\setlength{\tabcolsep}{0.5mm}{
\begin{table}[!htbp]

\caption{A comparison of the performances of different models on benchmark datasets. The first row shows the performance of ChatGPT using the manually designed prompt template.}\label{mainresult}
\centering
\begin{tabular}{l|c|c|c}
\hline 
\diagbox{Model}{F1(\%)}{Dataset}
& TB-Dense & TDDAuto & TDDMan \\ \hline
ChatGPT & 27.0 & - & 16.8   \\
\hline \hline
CAEVO\cite{Chambers14denseevent}  & 48.2     & 42.5    & 16.1   \\
EventPlus\cite{ma-etal-2021-eventplus} & 64.5     & 41.0    & 38.8   \\
CTRL-PG\cite{Zhou2021ClinicalTR}   & 65.2     & -       & -      \\
TIMERS\cite{mathur-etal-2021-timers}    & 67.8     & 71.1    & 45.5   \\
SCS-EERE\cite{TrongSOCS}  & -        & 76.7    & 51.1   \\
Unified-Framework\cite{huang-etal-2023-classification} & 68.1     & -    & -   \\
RSGT\cite{zhou-etal-2022-rsgt}      & 68.7     & -    & -   \\
\hline \hline
PETR(Ours) & \textbf{69.3} & \textbf{77.2}   & \textbf{52.6}      \\
\hline
\end{tabular}
\end{table}
}

Our conclusions are as follows: (1) According to the experimental results of Chan et al.,\cite{Chan2023ChatGPTEO} the extraction ability of ChatGPT\footnote{https://chat.openai.com} for TempRel task is limited. (2) The experimental results demonstrate that our PETR model outperforms existing baselines. (3) It proves the feasibility of using RAG on TempRel tasks.

%%%%%%%%%%%%%%%%%%%%%%%%%%%%%%%%%%%%%%%%%%%%%%%%%%%%%%%%%%%%%%%%%%%
\subsection{Performance on Individual Labels}
%The specific analysis is as follows:
\setlength{\tabcolsep}{1.5mm}{
    \begin{table*}[!htbp]
    \centering
    \caption{Model performance breakdown on the TDDiscourse datasets. }\label{tableragtdd}
\begin{tabular}{|l|llllll|llllll|}
\hline
\diagbox{Situation}{Dataset}             & \multicolumn{6}{c|}{TDDMan}                                                                                                                                                   & \multicolumn{6}{c|}{TDDAuto}                                                                                                                                                           \\ \hline
 Compare            & \multicolumn{3}{c|}{Without  Retrieval}                                                 & \multicolumn{3}{c|}{With  Retrieval}                                                & \multicolumn{3}{c|}{Without  Retrieval}                                                          & \multicolumn{3}{c|}{With  Retrieval}                                                \\ \hline
Metric       & \multicolumn{1}{c|}{P.}      & \multicolumn{1}{c|}{R.}      & \multicolumn{1}{c|}{F1.}     & \multicolumn{1}{c|}{P.}      & \multicolumn{1}{c|}{R.}      & \multicolumn{1}{c|}{F1.} & \multicolumn{1}{c|}{P.}      & \multicolumn{1}{c|}{R.}      & \multicolumn{1}{c|}{F1.}              & \multicolumn{1}{c|}{P.}      & \multicolumn{1}{c|}{R.}      & \multicolumn{1}{c|}{F1.} \\ \hline
BEFORE       & \multicolumn{1}{l|}{0.4879} & \multicolumn{1}{l|}{0.4514} & \multicolumn{1}{l|}{0.4689} & \multicolumn{1}{l|}{0.5113} & \multicolumn{1}{l|}{0.5062} & \textbf{0.5088}         & \multicolumn{1}{l|}{0.7682} & \multicolumn{1}{l|}{0.8758} & \multicolumn{1}{l|}{0.8185}          & \multicolumn{1}{l|}{0.7619} & \multicolumn{1}{l|}{0.9179} & \textbf{0.8327}         \\ \hline
AFTER        & \multicolumn{1}{l|}{0.3533} & \multicolumn{1}{l|}{0.2819} & \multicolumn{1}{l|}{0.3136} & \multicolumn{1}{l|}{0.3901} & \multicolumn{1}{l|}{0.3777} & \textbf{0.3838}         & \multicolumn{1}{l|}{0.8366} & \multicolumn{1}{l|}{0.8689} & \multicolumn{1}{l|}{0.8528}          & \multicolumn{1}{l|}{0.8557} & \multicolumn{1}{l|}{0.8780}  & \textbf{0.8667}         \\ \hline
SIMULTANEOUS & \multicolumn{1}{l|}{0.4151} & \multicolumn{1}{l|}{0.4889} & \multicolumn{1}{l|}{0.4490}  & \multicolumn{1}{l|}{0.4762} & \multicolumn{1}{l|}{0.4444} & \textbf{0.4598}         & \multicolumn{1}{l|}{0.7540}  & \multicolumn{1}{l|}{0.6894} & \multicolumn{1}{l|}{0.7202}          & \multicolumn{1}{l|}{0.7789} & \multicolumn{1}{l|}{0.7518} & \textbf{0.7651}         \\ \hline
INCLUDES     & \multicolumn{1}{l|}{0.5733} & \multicolumn{1}{l|}{0.5332} & \multicolumn{1}{l|}{0.5525} & \multicolumn{1}{l|}{0.6075} & \multicolumn{1}{l|}{0.5682} & \textbf{0.5872}         & \multicolumn{1}{l|}{0.6921} & \multicolumn{1}{l|}{0.5216} & \multicolumn{1}{l|}{\textbf{0.5948}} & \multicolumn{1}{l|}{0.7090}  & \multicolumn{1}{l|}{0.4702} & 0.5654                  \\ \hline
IS\_INCLUDED & \multicolumn{1}{l|}{0.4427} & \multicolumn{1}{l|}{0.5939} & \multicolumn{1}{l|}{0.5073} & \multicolumn{1}{l|}{0.4927} & \multicolumn{1}{l|}{0.5768} & \textbf{0.5314}         & \multicolumn{1}{l|}{0.5352} & \multicolumn{1}{l|}{0.4718} & \multicolumn{1}{l|}{0.5015}          & \multicolumn{1}{l|}{0.5974} & \multicolumn{1}{l|}{0.4524} & \textbf{0.5149}         \\ \hline
Overrall     & \multicolumn{3}{c|}{0.4903}                                                             & \multicolumn{3}{c|}{\textbf{0.5257}}                                                & \multicolumn{3}{c|}{0.7545}                                                                      & \multicolumn{3}{c|}{\textbf{0.7724}}                                                \\ \hline
\end{tabular}
\end{table*}
}
The specific analysis is as follows: As shown in Table \ref{tableragtdd}, after using RAG technology for PVP design, our model achieved significant performance improvements on most of the individual labels in the TDD-Man and TDD-Auto datasets.

\setlength{\tabcolsep}{2mm}{

\begin{table}[]\caption{Model performance breakdown on TB-Dense dataset.}\label{tablebd}
\centering
\begin{tabular}{|c|ccc|ccc|}
\hline
\diagbox{Situation}{F1(\%)}{Model}          & \multicolumn{3}{c|}{CTRL-PG}                                 & \multicolumn{3}{c|}{PETR(Ours)}                                    \\ \hline
metric       & \multicolumn{1}{c|}{P}    & \multicolumn{1}{c|}{R}    & F1   & \multicolumn{1}{c|}{P}    & \multicolumn{1}{c|}{R}    & F1   \\ \hline
Before       & \multicolumn{1}{c|}{52.6} & \multicolumn{1}{c|}{74.8} & 61.7 & \multicolumn{1}{c|}{75.9} & \multicolumn{1}{c|}{72.9} & \textbf{74.4} \\
After        & \multicolumn{1}{c|}{69.0}   & \multicolumn{1}{c|}{72.5} & 70.7 & \multicolumn{1}{c|}{86.2} & \multicolumn{1}{c|}{61.3} & \textbf{71.6} \\
Includes     & \multicolumn{1}{c|}{60.9} & \multicolumn{1}{c|}{29.8} & \textbf{40.0}   & \multicolumn{1}{c|}{41.0} & \multicolumn{1}{c|}{28.6} & 33.7 \\
Is\_Include   & \multicolumn{1}{c|}{34.7} & \multicolumn{1}{c|}{27.7} & 30.8 & \multicolumn{1}{c|}{43.8} & \multicolumn{1}{c|}{39.6}   & \textbf{41.6}   \\
Simultaneous & \multicolumn{1}{c|}{-}    & \multicolumn{1}{c|}{-}    & -    & \multicolumn{1}{c|}{54.6} & \multicolumn{1}{c|}{27.3} & \textbf{36.4} \\
Vague        & \multicolumn{1}{c|}{72.8} & \multicolumn{1}{c|}{64.8} & 68.6 & \multicolumn{1}{c|}{65.1} & \multicolumn{1}{c|}{78.1} & \textbf{71.0} \\ \hline
\end{tabular}

\end{table}
}

From Table \ref{tablebd}, we can see that our model has achieved significant improvements on most of the labels compared with the baseline model CTRL-PG\cite{Zhou2021ClinicalTR} on the TB-Dense dataset. In the Before label, our model has increased the total score of F1 by 12.7 \%. For the After label, it has an effective improvement of 0.9\% in F1. For Is\_Include label, our model has improved the performance by 10.8 \%. For the Vague label, our model has a 2.4\% improvement in F1 value. In terms of the SIMULTANEOUS label, referring to the relevant literature in recent years\cite{ma-etal-2021-eventplus, Zhou2021ClinicalTR, KJETE}, our model has made a historical breakthrough, with the F1 value increasing from 0 to 36.4\%. For the Includes label, the extraction effect of our method is 6.3\% lower than that of the CTRL-PG model, which shows that the prompt-based method cannot solve all the few-shot label problems. We will further explore this problem in the follow-up experiments.

\subsection{PVP Analysis}
%\begin{figure}[!htbp]

\begin{figure}[]
  \centering
  \includegraphics[width=.38\textwidth]{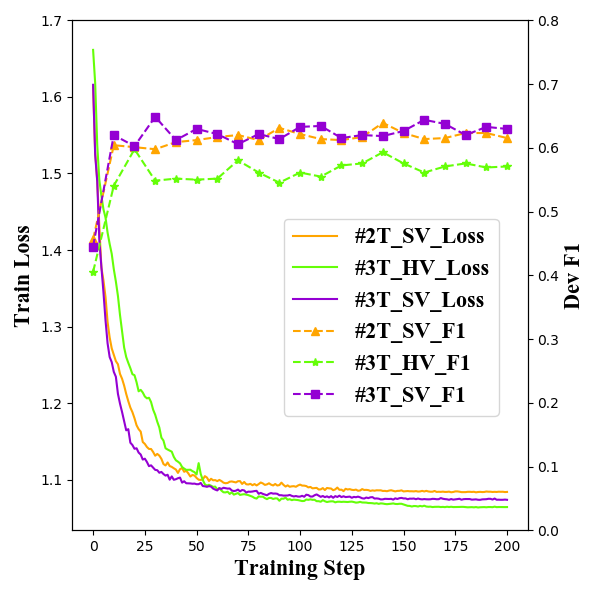}
  \caption{
  Convergence of different pvp pairs and their F1 scores in the validation dataset. The \#n T denotes the \#n Template in the Table.\ref{preliminaryExp}, and the SV, HV are the abbreviations of Soft Verbalizer, and Hard Verbalizer respectively. The \#3T\_SV (\#3 Template and Soft Verbalizer in the Table.\ref{preliminaryExp}) method converges faster than other methods, and it has a higher F1 score and an appropriate loss. This experiment demonstrates the importance of carefully choosing pvps.}\label{lossandprec}
\end{figure}

As is shown in Figure \ref{lossandprec}. We analyzed the dimensions of the template and verbalizer respectively. In the template dimension, the trigger or event is placed into the modifier word place in the post-decorated style template when they use the same soft verbalizer\cite{warp}. In the verbalizer dimension, we compare hard and soft verbalizers when the template is the same. Through experiments, it is found that the trigger has a better performance than the event. The soft verbalizer method is better than the hard method. We compare and analyze three experiments to verify that template and verbalizer design has a great impact on experimental results.

\subsection{Tuning Mode}

\begin{figure}[!htbp]
  \centering
  \includegraphics[width=.5\textwidth]{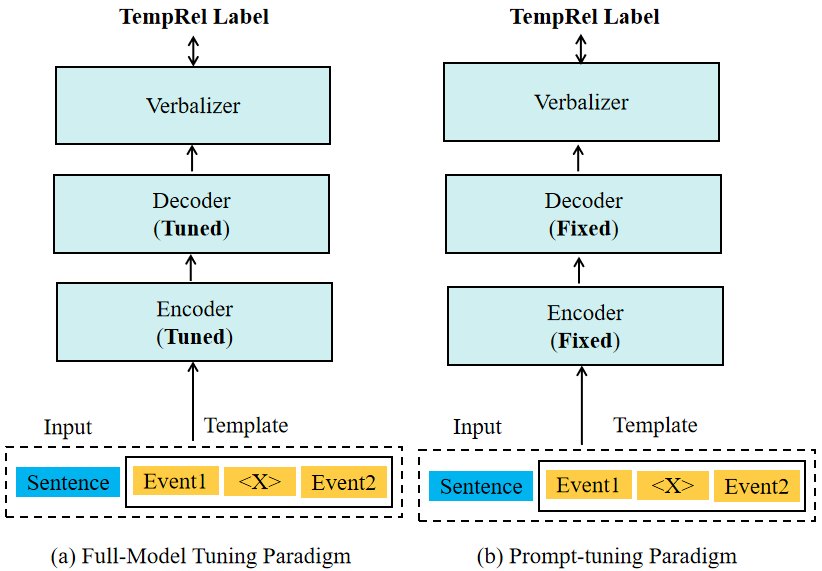}
  \caption{Paradigms of full-model tuning and prompt tuning. The $\langle X \rangle$ represents the traditional typical pre-training encoder-decoder mask model.}\label{figurepm}
\end{figure}

As Figure \ref{figurepm} shows, we used two model tuning methods, Full-Model Tuning (FT) and Prompt Tuning (PT). FT mode trains the entire model jointly with the soft prompt. In PT mode, only prompt tokens are tunable. In our experiment, we use the \#1 Template and Soft Verbalizer (in Table.\ref{preliminaryExp}) on the TB-Dense dataset to observe the differences between PT and FT methods (The same experimental phenomenon is observed using other PVPs). As shown in Table~\ref{tablepf}, the experimental results confirmed the following viewpoint: when the training data is sufficient, the extraction results of PT and FT are comparable \cite{gu-etal-2022-ppt, Ding2022OpenPromptAO}. However, when there is a lack of training data, FT outperforms its rival.

\setlength{\tabcolsep}{1.5mm}{
    \begin{table}[!h]
    \caption{Performance of learning under two different tuning modes.}\label{tablepf}
    \centering
    \begin{tabular}{lcc}
    \hline
          Mode            & \multicolumn{1}{l}{F1} & GPU Memory \\ \hline
    Prompt Tuning     & 51.72\%                      & 4610MiB    \\ \hline
    Full-Model Tuning & 66.96\%                      & 12461MiB   \\ \hline
    \end{tabular}
    
    \end{table}
}

\subsection{Comparison against baselines without retrieval}
As Table \ref{tablecmpllms} shows, we can draw the following conclusion: (1) We can observe that the performance of templates designed by auxiliary LLMs generally performs better than the manually designed ones, which proves the effectiveness of applying LLMs to the task of template design. (2) In most cases, soft verbalizers perform better than hard verbalizers, but there are exceptions (e.g., TDD-Auto) where hard verbalizers perform slightly better than soft verbalizers, further demonstrating the necessity for utilizing LLMs to assist in the design of verbalizers. (3) Experiments have shown that despite obtaining the best performance requires combining manual-design and LLM-design, the LLM still proves to be an effective tool for inspiring PVP design.

\setlength{\tabcolsep}{1.5mm}{
    \begin{table}[!htbp]\caption{Model performance breakdown on LLMs. Soft and hard represent the soft type of verbalizer and hard type of verbalizer, respectively. }\label{tablecmpllms}
    
    \centering
\begin{tabular}{lllllll}
\hline
\multicolumn{1}{l|}{\diagbox{Method}{Dataset}}            & \multicolumn{2}{c|}{TB-Dense}                           & \multicolumn{2}{c|}{TDDMan}                             & \multicolumn{2}{c}{TDDAuto}        \\ \hline 
\multicolumn{1}{c|}{}            & \multicolumn{1}{l|}{soft}  & \multicolumn{1}{c|}{hard}  & \multicolumn{1}{c|}{soft}  & \multicolumn{1}{c|}{hard}  & \multicolumn{1}{c|}{soft}  & hard  \\ \hline \hline
                                 & \multicolumn{6}{c}{Without  Retrieval}                                                                                                                 \\ \hline
\multicolumn{1}{c|}{Manual}      & \multicolumn{1}{l|}{\textbf{69.31}} & \multicolumn{1}{c|}{64.40}  & \multicolumn{1}{l|}{49.03} & \multicolumn{1}{l|}{43.96} & \multicolumn{1}{l|}{75.45} & 72.05 \\ \hline \hline
                                 & \multicolumn{6}{c}{With  Retrieval}                                                                                                                    \\ \hline
\multicolumn{1}{c|}{Erniebot3.5} & \multicolumn{1}{l|}{68.89} & \multicolumn{1}{c|}{67.13} & \multicolumn{1}{l|}{46.83} & \multicolumn{1}{l|}{46.63} & \multicolumn{1}{l|}{73.22} & 75.48 \\ \hline
\multicolumn{1}{c|}{LLama2-7B}   & \multicolumn{1}{l|}{68.82} & \multicolumn{1}{c|}{68.54} & \multicolumn{1}{l|}{49.90}  & \multicolumn{1}{l|}{43.36} & \multicolumn{1}{l|}{76.27} & 76.09 \\ \hline
\multicolumn{1}{c|}{Qwen-7B}     & \multicolumn{1}{l|}{68.04} & \multicolumn{1}{c|}{67.62} & \multicolumn{1}{l|}{50.63} & \multicolumn{1}{l|}{44.83} & \multicolumn{1}{l|}{76.39} & 76.09 \\ \hline
\multicolumn{1}{c|}{Taichu}      & \multicolumn{1}{l|}{68.68} & \multicolumn{1}{c|}{67.06} & \multicolumn{1}{l|}{\textbf{52.57}} & \multicolumn{1}{l|}{44.62} & \multicolumn{1}{l|}{75.69} & 75.97 \\ \hline
\multicolumn{1}{c|}{SparkDesk}   & \multicolumn{1}{l|}{68.04} & \multicolumn{1}{c|}{67.62} & \multicolumn{1}{l|}{48.83} & \multicolumn{1}{l|}{38.83} & \multicolumn{1}{l|}{75.69} & \textbf{77.24} \\ \hline
\multicolumn{1}{c|}{ChatGLM2-6B} & \multicolumn{1}{l|}{68.04} & \multicolumn{1}{c|}{67.20}  & \multicolumn{1}{l|}{50.63} & \multicolumn{1}{l|}{43.90}  & \multicolumn{1}{l|}{76.39} & 76.37 \\ \hline
\multicolumn{1}{c|}{GPT4}        & \multicolumn{1}{l|}{67.97} & \multicolumn{1}{c|}{65.80}  & \multicolumn{1}{l|}{51.43} & \multicolumn{1}{l|}{44.83} & \multicolumn{1}{l|}{75.69} & 74.58 \\ \hline
\end{tabular}
\end{table}
}

\subsection{The word frequency information returned by LLMs}
%\begin{figure}[!htbp]
\begin{figure}[!htbp]
  \centering
  \includegraphics[width=.5\textwidth]{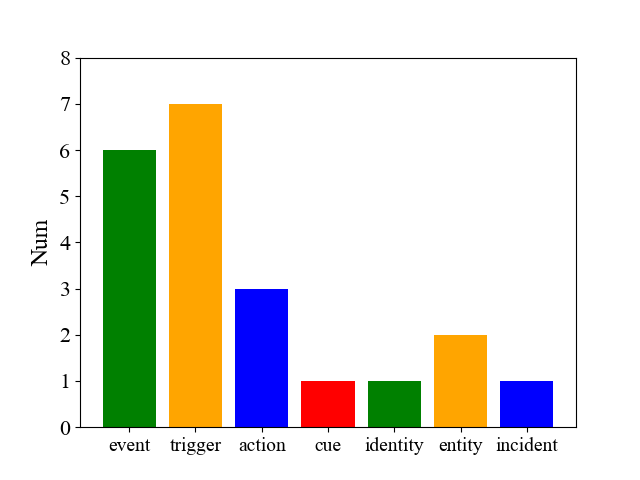}
  \caption{Modifier word frequency statistics returned by LLMs.}\label{templatecount1}
\end{figure}

Taking trigger modifier words as an example, we have counted the word frequency of modifier words returned by LLMs, as shown in Figure \ref{templatecount1}. Similar to human cognition, trigger-related words account for the highest proportion, followed by event, action, and other less apparent modifier words such as cue and identity.

\subsection{Correlation analysis}
To investigate the contribution of each module in the model, we conducted various relevant experiments. The following experiments are based on the template and verbalizer that achieved the best results on the TB-Dense dataset.

%\subsubsection{Masked v.s. Autoregressive v.s. Seq2seq Language Model}
\subsubsection{PLM selection}
%h
\begin{figure}[!htbp]
  \centering
  \includegraphics[width=.5\textwidth]{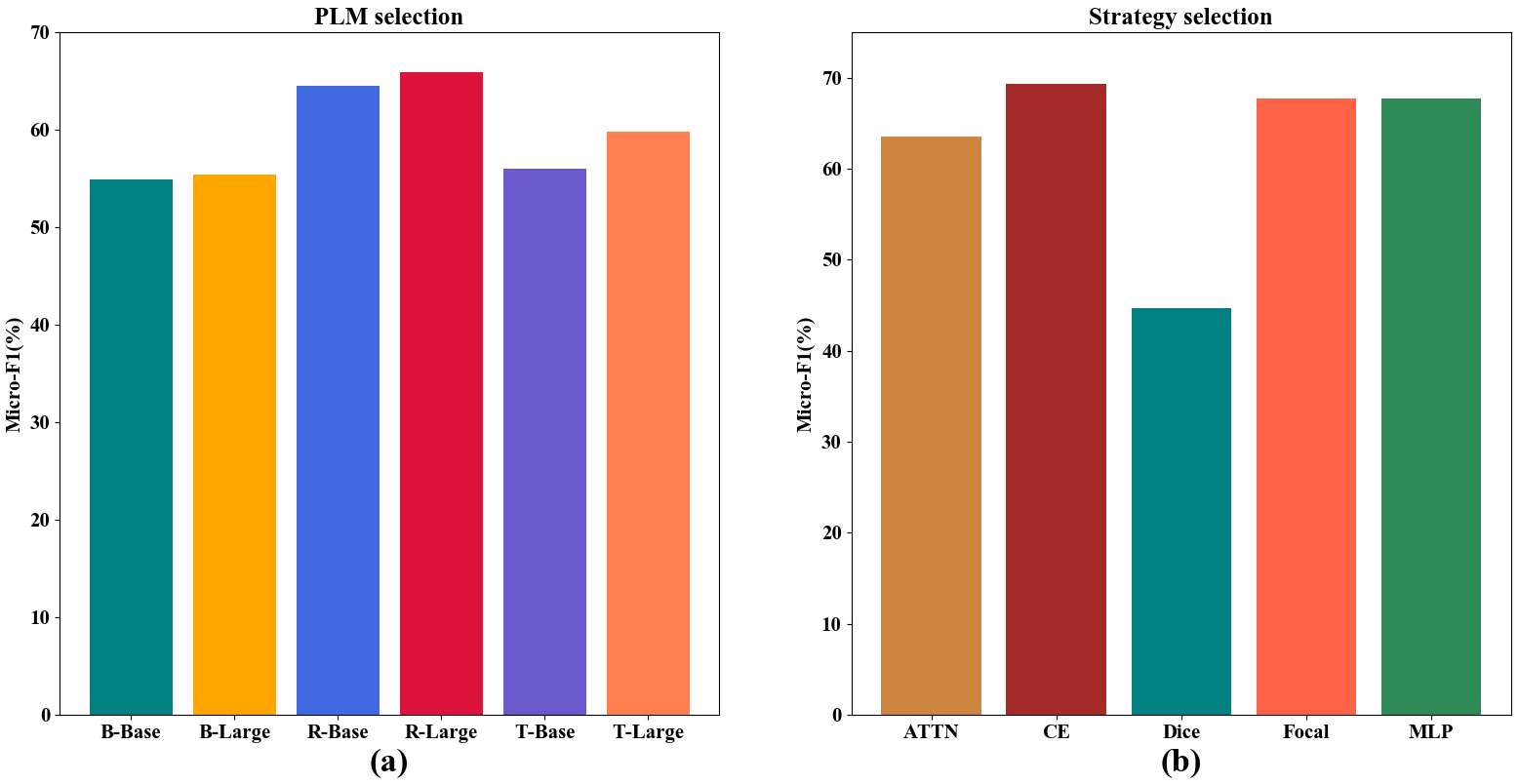}
  \caption{PLM and loss function selection, Where B denotes BERT, R denotes RoBERTa, T denotes T5, and G denotes GPT. Since RoBERTa-large outperforms the others, we use RoBERTa-large in the follow-up experiments. The Focal, Dice, CE, ATTN, and MLP denote FocalLoss, Dice Loss, Cross-Entropy,self-attention, and Multi-layer Perceptron, respectively.}\label{figureplm}
\end{figure}

As is shown in Fig.\ref{figureplm} (a), the performance of our model is sensitive to the PLMs. Among all PLMs, RoBERTa-Large yields the best performance. It is explainable since BERT and RoBERTa are MLMs that are good at dealing with discriminative tasks. Meanwhile, T5 and GPT are experts in generative tasks, so they did not perform as well as the MLMs in the case of TempRel extraction.

\iffalse
%\begin{figure}[!htbp]
\begin{figure}[!htbp]
  \centering
  \includegraphics[width=.3\textwidth]{./images/discussion3.jpg}
  \caption{The Focal, Dice, CE, ATTN, MLP denotes FocalLoss, Dice Loss, Cross-Entropy, self-attention, Multi-layer Perceptron, respectively.}\label{tablediscussion}
\end{figure}
\fi

\subsubsection{Strategy selection}
As is shown in Fig.\ref{figureplm} (b), After the selection of appropriate pre-trained language models. we introduced design strategies based on self-attention, FocalLoss\cite{Lin2017FocalLF}, DiceLoss\cite{DiceLoss}, and MLP to further improve the model performance, we put the self-attention or MLP behind the prompt model output layer and before the softmax layer, hoping to better capture temporal knowledge. The purpose of using the FocalLoss or DiceLoss is to introduce the specific loss function and expect the model to focus more on learning with few-shot samples. But through experiments, we found that the effect is good enough by directly passing through the prompt layer and then through the softmax layer. The traditional cross-entropy loss is the optimal choice, and there is no need to introduce an additional loss.

\subsection{Case Study}
\begin{figure}[!htbp]
  \centering
  \includegraphics[width=.46\textwidth]{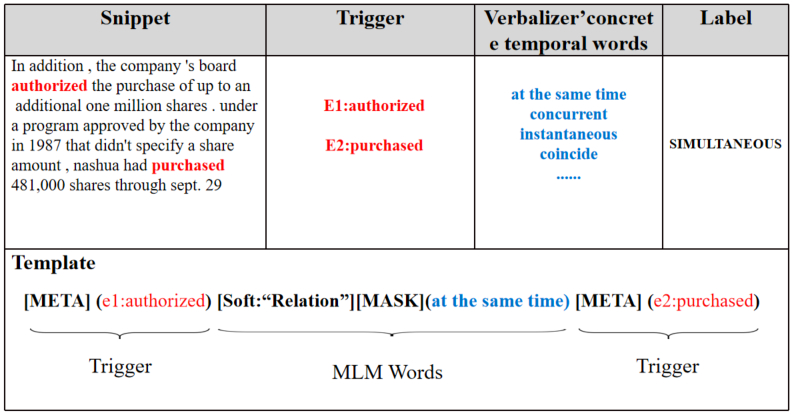}
  \caption{A illustration of prompt components}\label{tablecs}
\end{figure}

In Figure \ref{tablecs}, we show a case in the dataset to illustrate how our model utilizes the PLM's knowledge to help predict the right event temporal relation. We highlight the trigger word in the text. We can see that the trigger words are distributed in different clauses. Trigger words and TempRel relation are spelled into the model template. The concrete temporal words in the verbalizer component are a group of widely used temporal phrases. Let the model automatically select the appropriate word to fill the [MASK] position. We can see that the label and verbalizer’s concrete temporal words have a one-to-many relation. This design helps us take advantage of the potential temporal relation in the PLMs model. It can transform event temporal relation extraction tasks into cloze-style problems that PLM models are good at. 
%\begin{figure}[!htbp]

%%%%%%%%%%%%%%%%%%%%%%%%%%%%%%%%%%%%%%%%%%%%%%%%%%%%%%%%%%%%%%%%%%%
\section{Conclusion}
%%%%%%%%%%%%%%%%%%%%%%%%%%%%%%%%%%%%%%%%%%%%%%%%%%%%%%%%%%%%%%%%%%%
In this paper, we proposed a novel RAG-based approach for enhancing the design of PVPs within the TempRel extractio task, situated in the broader context of prompt learning paradigm. The primary goal of our proposed method is to identify the most suitable template and verbalizer to optimize TempRel extraction with the help of LLMs. Our approach includes two phases. In the first phase, we prompted mainstream LLMs to generate a diverse array of candidate PVPs, crafted for the TempRel task. In the second phase, we developed and implemented an algorithm to identify the most effective PVP for each specific TempRel dataset. To validate the efficacy of our method, we conducted comprehensive comparative experiments which spanned across various dimensions, including overall performance, individual label performance, and different tuning modes. The experimental results on three widely used datasets confirm that our approach is effective for improving the performance of TempRel extraction.

\vspace{12pt}
%\color{red}


\begin{thebibliography}{00}

\bibitem{han2021ptr}
X.~Han, W.~Zhao, N.~Ding, Z.~Liu, and M.~Sun, ``Ptr: Prompt tuning with rules for text classification,'' {\em arXiv preprint arXiv:2105.11259}, 2021.

\iffalse
\bibitem{Jiang2023StructGPTAG}
J.~Jiang, K.~Zhou, Z.~Dong, K.~Ye, W.~X. Zhao, and J.~rong Wen, ``Structgpt: A general framework for large language model to reason over structured data,'' {\em ArXiv}, vol.~abs/2305.09645, 2023.
\fi

\bibitem{Touvron2023Llama2O}
H.~Touvron, L.~Martin, K.~R. Stone, P.~Albert, A.~Almahairi, Y.~Babaei, N.~Bashlykov, S.~Batra, P.~Bhargava, S.~Bhosale, D.~M. Bikel, L.~Blecher, C.~C. Ferrer, M.~Chen, G.~Cucurull, D.~Esiobu, J.~Fernandes, J.~Fu, W.~Fu, B.~Fuller, C.~Gao, V.~Goswami, N.~Goyal, A.~S. Hartshorn, S.~Hosseini, R.~Hou, H.~Inan, M.~Kardas, V.~Kerkez, M.~Khabsa, I.~M. Kloumann, A.~V. Korenev, P.~S. Koura, M.-A. Lachaux, T.~Lavril, J.~Lee, D.~Liskovich, Y.~Lu, Y.~Mao, X.~Martinet, T.~Mihaylov, P.~Mishra, I.~Molybog, Y.~Nie, A.~Poulton, J.~Reizenstein, R.~Rungta, K.~Saladi, A.~Schelten, R.~Silva, E.~M. Smith, R.~Subramanian, X.~Tan, B.~Tang, R.~Taylor, A.~Williams, J.~X. Kuan, P.~Xu, Z.~Yan, I.~Zarov, Y.~Zhang, A.~Fan, M.~Kambadur, S.~Narang, A.~Rodriguez, R.~Stojnic, S.~Edunov, and T.~Scialom, ``Llama 2: Open foundation and fine-tuned chat models,'' {\em ArXiv}, vol.~abs/2307.09288, 2023.

\bibitem{huang-etal-2023-classification}
Q.~Huang, Y.~Hu, S.~Zhu, Y.~Feng, C.~Liu, and D.~Zhao, ``More than classification: A unified framework for event temporal relation extraction,'' in {\em Proceedings of the 61st Annual Meeting of the Association for Computational Linguistics (Volume 1: Long Papers)}, (Toronto, Canada), pp.~9631--9646, Association for Computational Linguistics, July 2023.

\bibitem{DBLP:conf/iconip/WangYM23}
Z.~Wang, Y.~Yang, and J.~Ma, ``Two-stage graph convolutional networks for relation extraction,'' in {\em Neural Information Processing - 30th International Conference, {ICONIP} 2023, Changsha, China, November 20-23, 2023, Proceedings, Part {XV}} (B.~Luo, L.~Cheng, Z.~Wu, H.~Li, and C.~Li, eds.), vol.~1969 of {\em Communications in Computer and Information Science}, pp.~483--494, Springer, 2023.

\bibitem{KJETE}
X.~Zhang, L.~Zang, P.~Cheng, Y.~Wang, and S.~Hu, ``A knowledge/data enhanced method for joint event and temporal relation extraction,'' in {\em ICASSP 2022 - 2022 IEEE International Conference on Acoustics, Speech and Signal Processing (ICASSP)}, pp.~6362--6366, 2022.

\bibitem{TrongSOCS}
H.~Man, N.~T. Ngo, L.~N. Van, and T.~H. Nguyen, ``Selecting optimal context sentences for event-event relation extraction,'' in {\em AAAI Conference on Artificial Intelligencel Intelligence}, 2022.

\bibitem{han-etal-2021-econet}
R.~Han, X.~Ren, and N.~Peng, ``{ECONET}: Effective continual pretraining of language models for event temporal reasoning,'' in {\em Proceedings of the 2021 Conference on Empirical Methods in Natural Language Processing}, (Online and Punta Cana, Dominican Republic), pp.~5367--5380, Association for Computational Linguistics, Nov. 2021.

\bibitem{Zhou2021ClinicalTR}
Y.~Zhou, Y.~Yan, R.~Han, J.~H. Caufield, K.-W. Chang, Y.~Sun, P.~Ping, and W.~Wang, ``Clinical temporal relation extraction with probabilistic soft logic regularization and global inference,'' in {\em AAAI Conference on Artificial Intelligencel Intelligence}, 2021.

\bibitem{cheng-miyao-2017-classifying}
F.~Cheng and Y.~Miyao, ``Classifying temporal relations by bidirectional {LSTM} over dependency paths,'' in {\em Proceedings of the 55th Annual Meeting of the Association for Computational Linguistics (Volume 2: Short Papers)}, (Vancouver, Canada), pp.~1--6, Association for Computational Linguistics, July 2017.

\bibitem{mathur-etal-2021-timers}
P.~Mathur, R.~Jain, F.~Dernoncourt, V.~Morariu, Q.~H. Tran, and D.~Manocha, ``{TIMERS}: Document-level temporal relation extraction,'' in {\em Proceedings of the 59th Annual Meeting of the Association for Computational Linguistics and the 11th International Joint Conference on Natural Language Processing (Volume 2: Short Papers)}, (Online), pp.~524--533, Association for Computational Linguistics, Aug. 2021.

\bibitem{Chambers14denseevent}
N.~Chambers, T.~Cassidy, B.~McDowell, and S.~Bethard, ``Dense event ordering with a multi-pass architecture,'' 2014.

\bibitem{cassidy-EtAl:2014:P14-2}
T.~Cassidy, B.~McDowell, N.~Chambers, and S.~Bethard, ``An annotation framework for dense event ordering,'' in {\em Proceedings of the 52nd Annual Meeting of the Association for Computational Linguistics (Volume 2: Short Papers)}, (Baltimore, Maryland), pp.~501--506, Association for Computational Linguistics, June 2014.

\bibitem{naik-etal-2019-tddiscourse}
A.~Naik, L.~Breitfeller, and C.~Rose, ``{TDD}iscourse: A dataset for discourse-level temporal ordering of events,'' in {\em Proceedings of the 20th Annual SIGdial Meeting on Discourse and Dialogue}, (Stockholm, Sweden), pp.~239--249, Association for Computational Linguistics, Sept. 2019.

\bibitem{2003TB}
J.~Pustejovsky, P.~Hanks, R.~Saurí, A.~See, and M.~Lazo, ``The timebank corpus,'' {\em proceedings of corpus linguistics}, 2003.

\bibitem{ning-etal-2019-improved}
Q.~Ning, S.~Subramanian, and D.~Roth, ``An improved neural baseline for temporal relation extraction,'' in {\em Proceedings of the 2019 Conference on Empirical Methods in Natural Language Processing and the 9th International Joint Conference on Natural Language Processing (EMNLP-IJCNLP)}, (Hong Kong, China), pp.~6203--6209, Association for Computational Linguistics, Nov. 2019.

\bibitem{ning-etal-2018-improving}
Q.~Ning, H.~Wu, H.~Peng, and D.~Roth, ``Improving temporal relation extraction with a globally acquired statistical resource,'' in {\em Proceedings of the 2018 Conference of the North {A}merican Chapter of the Association for Computational Linguistics: Human Language Technologies, Volume 1 (Long Papers)}, (New Orleans, Louisiana), pp.~841--851, Association for Computational Linguistics, June 2018.

\bibitem{Lewis2020RetrievalAugmentedGF}
P.~Lewis, E.~Perez, A.~Piktus, F.~Petroni, V.~Karpukhin, N.~Goyal, H.~Kuttler, M.~Lewis, W.~tau Yih, T.~Rockt{\"a}schel, S.~Riedel, and D.~Kiela, ``Retrieval-augmented generation for knowledge-intensive nlp tasks,'' {\em ArXiv}, vol.~abs/2005.11401, 2020.

\bibitem{Chen2021KnowPromptKP}
X.~Chen, N.~Zhang, N.~Zhang, X.~Xie, S.~Deng, Y.~Yao, C.~Tan, F.~Huang, L.~Si, and H.~Chen, ``Knowprompt: Knowledge-aware prompt-tuning with synergistic optimization for relation extraction,'' {\em Proceedings of the ACM Web Conference 2022}, 2021.

\bibitem{hu-etal-2022-knowledgeable}
S.~Hu, N.~Ding, H.~Wang, Z.~Liu, J.~Wang, J.~Li, W.~Wu, and M.~Sun, ``Knowledgeable prompt-tuning: Incorporating knowledge into prompt verbalizer for text classification,'' in {\em Proceedings of the 60th Annual Meeting of the Association for Computational Linguistics (Volume 1: Long Papers)} (S.~Muresan, P.~Nakov, and A.~Villavicencio, eds.), (Dublin, Ireland), pp.~2225--2240, Association for Computational Linguistics, May 2022.

\bibitem{gu-etal-2022-ppt}
Y.~Gu, X.~Han, Z.~Liu, and M.~Huang, ``{PPT}: Pre-trained prompt tuning for few-shot learning,'' in {\em Proceedings of the 60th Annual Meeting of the Association for Computational Linguistics (Volume 1: Long Papers)}, (Dublin, Ireland), pp.~8410--8423, Association for Computational Linguistics, May 2022.

\bibitem{gao-etalLMBFF}
T.~Gao, A.~Fisch, and D.~Chen, ``Making pre-trained language models better few-shot learners,'' in {\em Proceedings of the 59th Annual Meeting of the Association for Computational Linguistics and the 11th International Joint Conference on Natural Language Processing (Volume 1: Long Papers)}, (Online), pp.~3816--3830, Association for Computational Linguistics, Aug. 2021.

\bibitem{warp}
K.~Hambardzumyan, H.~Khachatrian, and J.~May, ``{WARP}: {W}ord-level {A}dversarial {R}e{P}rogramming,'' in {\em Proceedings of the 59th Annual Meeting of the Association for Computational Linguistics and the 11th International Joint Conference on Natural Language Processing (Volume 1: Long Papers)}, (Online), pp.~4921--4933, Association for Computational Linguistics, Aug. 2021.

\bibitem{Chan2023ChatGPTEO}
C.~Chan, J.~Cheng, W.~Wang, Y.~Jiang, T.~Fang, X.~Liu, and Y.~Song, ``Chatgpt evaluation on sentence level relations: A focus on temporal, causal, and discourse relations,'' {\em ArXiv}, vol.~abs/2304.14827, 2023.

\bibitem{Robinson2022LeveragingLL}
J.~Robinson, C.~M. Rytting, and D.~Wingate, ``Leveraging large language models for multiple choice question answering,'' {\em ArXiv}, vol.~abs/2210.12353, 2022.

\bibitem{Ding2022OpenPromptAO}
N.~Ding, S.~Hu, W.~Zhao, Y.~Chen, Z.~Liu, H.~Zheng, and M.~Sun, ``{O}pen{P}rompt: An open-source framework for prompt-learning,'' in {\em Proceedings of the 60th Annual Meeting of the Association for Computational Linguistics: System Demonstrations}, (Dublin, Ireland), pp.~105--113, Association for Computational Linguistics, May 2022.

\bibitem{xu-etal-2023-retrieval}
B.~Xu, C.~Zhao, W.~Jiang, P.~Zhu, S.~Dai, C.~Pang, Z.~Sun, S.~Wang, and Y.~Sun, ``Retrieval-augmented domain adaptation of language models,'' in {\em Proceedings of the 8th Workshop on Representation Learning for NLP (RepL4NLP 2023)} (B.~Can, M.~Mozes, S.~Cahyawijaya, N.~Saphra, N.~Kassner, S.~Ravfogel, A.~Ravichander, C.~Zhao, I.~Augenstein, A.~Rogers, K.~Cho, E.~Grefenstette, and L.~Voita, eds.), (Toronto, Canada), pp.~54--64, Association for Computational Linguistics, July 2023.

\bibitem{Lin2017FocalLF}
T.-Y. Lin, P.~Goyal, R.~B. Girshick, K.~He, and P.~Doll{\'a}r, ``Focal loss for dense object detection,'' {\em 2017 IEEE International Conference on Computer Vision (ICCV)}, pp.~2999--3007, 2017.

\bibitem{DiceLoss}
F.~Milletari, N.~Navab, and S.-A. Ahmadi, ``V-net: Fully convolutional neural networks for volumetric medical image segmentation,'' in {\em 2016 Fourth International Conference on 3D Vision (3DV)}, pp.~565--571, 2016.


\bibitem{ma-etal-2021-eventplus}
M.~D. Ma, J.~Sun, M.~Yang, K.-H. Huang, N.~Wen, S.~Singh, R.~Han, and N.~Peng, ``{E}vent{P}lus: A temporal event understanding pipeline,'' in {\em Proceedings of the 2021 Conference of the North American Chapter of the Association for Computational Linguistics: Human Language Technologies: Demonstrations}, (Online), pp.~56--65, Association for Computational Linguistics, June 2021.

\bibitem{jiang-etal-2023-structgpt}
J.~Jiang, K.~Zhou, Z.~Dong, K.~Ye, X.~Zhao, and J.-R. Wen, ``{S}truct{GPT}: A general framework for large language model to reason over structured data,'' in {\em Proceedings of the 2023 Conference on Empirical Methods in Natural Language Processing} (H.~Bouamor, J.~Pino, and K.~Bali, eds.), (Singapore), pp.~9237--9251, Association for Computational Linguistics, Dec. 2023.

\bibitem{zhou-etal-2022-rsgt}
J.~Zhou, S.~Dong, H.~Tu, X.~Wang, and Y.~Dou, ``{RSGT}: Relational structure guided temporal relation extraction,'' in {\em Proceedings of the 29th International Conference on Computational Linguistics}, (Gyeongju, Republic of Korea), pp.~2001--2010, International Committee on Computational Linguistics, Oct. 2022.

\bibitem{yu-etal-2023-retrieval}
G.~Yu, L.~Liu, H.~Jiang, S.~Shi, and X.~Ao, ``Retrieval-augmented few-shot text classification,'' in {\em Findings of the Association for Computational Linguistics: EMNLP 2023} (H.~Bouamor, J.~Pino, and K.~Bali, eds.), (Singapore), pp.~6721--6735, Association for Computational Linguistics, Dec. 2023.

\bibitem{zhang-etal-2023-reaugkd}
J.~Zhang, A.~Muhamed, A.~Anantharaman, G.~Wang, C.~Chen, K.~Zhong, Q.~Cui, Y.~Xu, B.~Zeng, T.~Chilimbi, and Y.~Chen, ``{R}e{A}ug{KD}: Retrieval-augmented knowledge distillation for pre-trained language models,'' in {\em Proceedings of the 61st Annual Meeting of the Association for Computational Linguistics (Volume 2: Short Papers)} (A.~Rogers, J.~Boyd-Graber, and N.~Okazaki, eds.), (Toronto, Canada), pp.~1128--1136, Association for Computational Linguistics, July 2023.


\iffalse
\bibitem{b1} G. Eason, B. Noble, and I. N. Sneddon, ``On certain integrals of Lipschitz-Hankel type involving products of Bessel functions,'' Phil. Trans. Roy. Soc. London, vol. A247, pp. 529--551, April 1955.
\bibitem{b2} J. Clerk Maxwell, A Treatise on Electricity and Magnetism, 3rd ed., vol. 2. Oxford: Clarendon, 1892, pp.68--73.
\bibitem{b3} I. S. Jacobs and C. P. Bean, ``Fine particles, thin films and exchange anisotropy,'' in Magnetism, vol. III, G. T. Rado and H. Suhl, Eds. New York: Academic, 1963, pp. 271--350.
\bibitem{b4} K. Elissa, ``Title of paper if known,'' unpublished.
\bibitem{b5} R. Nicole, ``Title of paper with only first word capitalized,'' J. Name Stand. Abbrev., in press.
\bibitem{b6} Y. Yorozu, M. Hirano, K. Oka, and Y. Tagawa, ``Electron spectroscopy studies on magneto-optical media and plastic substrate interface,'' IEEE Transl. J. Magn. Japan, vol. 2, pp. 740--741, August 1987 [Digests 9th Annual Conf. Magnetics Japan, p. 301, 1982].
\bibitem{b7} M. Young, The Technical Writer's Handbook. Mill Valley, CA: University Science, 1989.
\fi

\end{thebibliography}
\end{document}